\documentclass[sn-mathphys]{sn-jnl}

\jyear{2021}%

\theoremstyle{thmstyleone}%
\usepackage{natbib}
\setcitestyle{numbers,square}
\theoremstyle{thmstyletwo}%

\theoremstyle{thmstylethree}%
\usepackage{subfigure}
\usepackage{hyperref}       
\usepackage{url}            
\usepackage{booktabs}       
\usepackage{amsfonts}       
\usepackage{nicefrac}       
\usepackage{microtype}      
\usepackage{amsmath}
\newtheorem{myDef}{Definition}
\newtheorem{mytheorem}{Theorem}
\newtheorem{myProblem}{Problem}
\newtheorem{myproposition}{Proposition}
\usepackage{graphicx}
\usepackage{mathrsfs}
\begin{document}

\title[Article Title]{Mixed-Precision Inference Quantization: Radically Towards Faster inference speed,  Lower Storage requirement, and Lower Loss }


\author[1]{\fnm{Cheng Daning}  }\email{chengdn@mail.tsinghua.edu.cn}

\author*[1]{\fnm{Chen Wenguang} }\email{cwg@mail.tsinghua.edu.cn}

\affil[1]{\orgdiv{Computer Science Department}, \orgname{Tsinghua University}, \orgaddress{ \city{Beijing}, \postcode{100080}, \state{Beijing}, \country{China}}}


\abstract{Based on the model's resilience to computational noise, model quantization is important for compressing models and improving computing speed. Existing quantization techniques rely heavily on experience and "fine-tuning" skills. In the majority of instances, the quantization model has a larger loss than a full precision model. This study provides a methodology for acquiring a mixed-precise quantization model with a lower loss than the full precision model. In addition, the analysis demonstrates that, throughout the inference process, the loss function is mostly affected by the noise of the layer inputs. In particular, we will demonstrate that neural networks with massive identity mappings are resistant to the quantization method. It is also difficult to improve the performance of these networks using quantization.}




\maketitle

{\section{Introduction}}
Neural network storage, inference, and training are computationally and temporally intensive due to the massive parameter size of neural networks, which needs a big memory footprint and a high number of floating-point operations per second (FLOPS). Therefore, developing a compression algorithm for machine learning models is necessary. Model quantization, based on the robustness of computational noise, is one of the most important compression techniques. The computational noise robustness measures the algorithm's performance when noise is added during the computation process.   The primary sources of noise are truncation and data type conversion mistakes. 

In quantization process, the initial high-precision data type used for a model's parameters is replaced with a lower-precision data type during model quantization. It is typical to replace FP32 with FP16 and both PyTorch and TensorFlow have quantization techniques that translate floats to integers. Various quantization techniques share the same theoretical foundation, which is the substitution of approximation data for the original data in the storage and inference processes. A lower-precision data format requires less memory, and using lower-precision data requires fewer computer resources and less time. In quantization, the precision loss in different quantization level conversions and data type conversions is the source of the noise.

The primary issue with the model quantization strategy is that a naive quantization scheme is likely to raise the loss function. It is not easy to substitute massive-scale model parameters with extremely low-data precision without sacrificing significant precision. It is not possible to utilize the same quantization level, i.e., to introduce the same level of noise to all parameters for all model parameters, and yet get good performance.

Utilizing mixed-precision quantization is one way to solve this issue. For more "sensitive" model parameters, higher-precision data is used, whereas lower-precision data is used for "nonsensitive" model parameters. Higher-precision data indicates that the original data adds minimal value noise, while lower-precision data indicates that the original data adds considerable value noise.

However, the consensus of academics is that quantization technology without "fine-tuning" is detrimental to the performance of the model. No study examines how to increase the performance of a model using quantization technology or which types of models are stable in the quantization process and why.

Moreover, the search space for a mixed-precision layout strategy is expansive. The following facts challenge current algorithms for determining the optimal mixed-precision layout strategy: These algorithms are built on empirical experience and "tuning" skills. Some forego neural network and dataset analysis and instead base model quantization on hardware features. It is impossible to set a precise limit for determining how various algorithms perform better under what circumstances. Some algorithms utilize Hessian data. The majority of them are analyzable. However, obtaining Hessian information necessitates a considerable amount of processing resources and time. Some of these methods are only useful for storing purposes.

This research provides a basic analysis of the computational noise robustness of neural networks. The model quantization problem setting for inference processes will be established. Moreover, by focusing on layerwise post-training static model quantization, we demonstrate that in the inference process, conversion loss due to varying quantization levels at the inputs to distinct layers plays a predominant role in loss functions. In addition, we present a method for acquiring a quantized model with a lower loss than the model with full precision. In addition, based on our analysis, we also prove that it is the nature that the computational noise robustness is strong for the neural network that mainly consists of identity mapping, like ResNet or DenseNet.
%
%
%
%

\section{Related Work}
Model compression methods include pruning methods\cite{han2015learning,li2016pruning,mao2017exploring} , knowledge distillation\cite{hinton2015distilling}, weight sharing\cite{ullrich2017soft} and quantization methods.  From the perspective of the precision layout, post-training quantization methods can be mainly divided into channelwise \cite{2019Fully,2020Channel}, groupwise \cite{dong2019hawq} and layerwise \cite{2019HAWQ} methods.   Layerwise mixed-precision layout schemes are more friendly to hardware. Parameters of the same precision are organized together, making full of a program's temporal and spatial locality. A common problem definition for quantization\cite{2019HAWQ,morgan1991experimental,courbariaux2015binaryconnect,2020HAWQV3}  is as follows \cite{gholami2021survey}.

\begin{myProblem}
	The objective of quantization is to solve the following optimization problem: \label{ori_pro_def}
	\begin{equation*}
		\min_{q \in \mathbf{Q}}\| q(w) - w\|^2
	\end{equation*}
	where $q$ is the quantization scheme, $q(w)$ is the quantized model with quantization $q$, and $w$ represents the weights, i.e., parameters, in the neural network.
\end{myProblem}

Quantization methods replace original data with lower-bit representations. Quantization methods can be divided into post-training quantization and quantization-aware training. This paper considers post-training quantization methods. Quantization produces a model with a small memory cost and a high arithmetic intensity without changing the structure of the original neural network. Most quantization methods are designed for mixed-precision quantization \cite{2019HAWQ,2020HAWQV3,dong2019hawq,wu2018mixed,wang2019haq,yu2020search}. In a mixed-precision layout scheme, some layers are stored at higher precision, while others are kept at a lower precision. However, a challenge that must be faced in this approach is how to find the correct mixed-precision settings for the different layers. A brute-force approach is not feasible since the search space is exponentially large in the number of layers.

Although problem \ref{ori_pro_def} gives researchers a target to aim for when performing quantization, the current problem definition has two shortcomings: 1. The search space of all possible mixed-precision layout schemes is a discrete space that is exponentially large in the number of layers. There is no effective method to solve the corresponding search problem. 2. There is a gap between the problem target and the final task target. As we can see, no terms related to the final task target, such as the loss function or accuracy, appear in the current problem definition.

\section{Background Analysis}
\subsection{Model Computation, Noise Generation and Quantization \label{ch:device cha}}

Compressed models for the inference process are computed using different methods depending on the hardware, programming methods and deep learning framework. All of these methods introduce noise into the computing process.

One reason for this noise problem is that although it is common practice to store and compute model parameters directly using different data types, only data of the same precision can support precise computations in a computer framework. 

Therefore, before performing computations on nonuniform data, a computer will convert them into the same data type. Usually, a lower-precision data type in a standard computing environment will be converted into a higher-precision data type; this ensures that the results are correct but require more computational resources and time. However, to accelerate the computing speed, some works on artificial intelligence (AI) computations propose converting higher-precision data types into lower-precision data types based on the premise that AI models are not sensitive to compression noise.  The commonly used quantization technology is converting data directly and using a lower-precision data type to map to a higher-precision data type linearly. 

We use the following example to illustrate quantization method, which is presented in \cite{2020HAWQV3}. Suppose that there are two data objects $input_1$ and $input_2$ are to be subjected to a computing operation, such as multiplication. After the quantization process, we have $Q_1 = \mbox{int}(\frac{input_1}{scale_1})$ and $Q_2 = \mbox{int}(\frac{input_2}{scale_2})$, and we can write
\begin{equation*}
	Q_{output}=\mbox{int}(\frac{input_1 * input_2}{scale_{output}}) \approx\mbox{int}( Q_1Q_2\frac{scale_1*scale_2}{scale_{output}})
\end{equation*}
$scale_{output}$, $scale_1$ and $scale_2$ are precalculated scale factors that depend on the distributions of $input_1$, $input_2$ and the output; $Q_i$ is stored as a lower-precision data type, such as an integer. All $scale$ terms can be precalculated and established ahead of time. Then, throughout the whole inference process, only computations on the $Q_i$ values are needed, which are fast. In this method, the noise is introduced in the $\mbox{int}(\cdot)$ process. This basic idea gives rise to several variants, such as (non)uniform quantization and (non)symmetrical quantization.

When we focus on quantization strategy, i.e. $round$ function in quantization framework like Micronet, we can have at least three strategy: round up, i.e., $ceil$ function in python, round down, i.e., $floor$ function in python and rounding, i.e., $round$ function in python. usually, rounding is the most common method to deal with quantization. But, in this paper, we will show that how to mixed use round up/round down to gain a mixed precision quantized model which is better than full precision model.
\subsection{Neural Networks}

In this paper, we mainly use the mathematical properties of extreme points to analyze quantization methods. This approach is universal to all cases, not only neural networks. However, there is a myth in the community that it is the neural network properties that guarantee the success of quantization methods\cite{wang2019haq,morgan1991experimental,2020Effective}. To show that the properties of the extreme points, not the properties of the neural network, are what determine the ability to quantize, i.e. the ability to handle noise, we must first define what a neural network is.

The traditional definition of a neural network  \cite{Denilson2016Understanding} as a human brain simulation is ambiguous; it is not a rigorous mathematical concept and cannot offer any analyzable information. The traditional descriptions of neural networks \cite{Denilson2016Understanding} focus on the inner products of the network weights and inputs, the activation functions and directed acyclic graphs. However, with the development of deep learning, although most neural networks still consist of weighted connections and activation layers, many neural networks no longer obey these rules, such as the network architectures for position embedding and layer norm operations in Transformers. Moreover, current deep learning frameworks, such as PyTorch and TensorFlow, offer application programming interfaces (APIs) to implement any function in a layer. Therefore, we propose that the definition of a neural network adheres to the engineering concept indicated by the definition \ref{concept nn} rather than a precise mathematical definition; that is, a neural network is a way for implementing a function.

\begin{myDef}
	The neural network is the function which is implemented in composite function form.
	\label{concept nn}
\end{myDef}

A neural network can be described in the following Eq. \ref{NN form} form.
\begin{align}
	model(x) = h_1(h_{2,1}(h_{3,1}(...),...,h_{3,k},w_{2,1}), \notag\\
	h_{2,2}(h_{3,k+1}(...),...,w_3),...,w_{2,2}),
	...,w_1) \label{NN form}
\end{align}
where  $h_{i,j}$, $i\in[2,...,n]$, are the ($n-i+1$)th layers in the neural network; $w_{i,j}$ is the parameter in $h_{i,j}(\cdot)$.

Definition \ref{concept nn} means that a neural network, without training, can be any function. With definition \ref{concept nn}, a neural network is no longer a mathematical concept, but this idea is widely used in practice \cite{2019Relay}. We can see from definition \ref{concept nn} that the requirement that a neural network is in composite function form is the only mathematical property of a neural network that can be used for analysis.

In practice, the loss function is one method to evaluate a neural network. A lower loss on a dataset means a better performance neural network. For example, the training process optimises the model's loss, i.e., following Eq. \ref{training problem}.
\begin{equation}
	\min_{w}	f(w) =\mathbf{E}_{sample} \ell(w,sample) =\frac{1}{m}\sum_{(x_i,y_i) \in \mathbb{D}} \ell(w,x_i,y_i)
	\label{training problem}
\end{equation}
where $f(\cdot)$ is the loss for model on a dataset, $w$ represents the model parameters, $\mathbb{D}$ is the dataset, $m$ is the size of the dataset, $\ell(\cdot)$ is the loss function for a sample and $(x_i,y_i)$ represents a sample in the dataset and its label. 

In this paper, we mainly use the sequential neural network to describe the conclusion for the sequential neural network is easily described, and the whole conclusion is non-related to the structure of the neural network. For a sequential $n$-layer neural network, $\ell(\cdot)$ can be described in the following Eq.\ref{seq nn} form.
\begin{align}
	&\ell(w,x_i,y_i) = L(model_n(x_i,w),y_i) \notag\\
	&model_n = h_1(h_2(h_3(h_4(\cdots h_n(h_{n+1},w_n) \notag\\
	&\cdots,w_4),w_3),w_2),w_1)
	\label{seq nn}
\end{align}
where $L(\cdot)$ is the loss function, such as the cross-entropy function; $h_i$, $i\in[1,...,n]$, is the ($n-i+1$)th layer in the neural network; $w = (w_n^T,w_{n-1}^T,\cdots,w_1^T)^T$, $w_i$ is the parameter in $h_i(\cdot)$; and for a unified format, $h_{n+1}$ stands for the sample $x$.

\section{Analysis of Computational Noise Robustness and Quantization for Inference}
\subsection{Start point}
\subsubsection{Analysis Base}
\begin{figure}[!htbp]
	\centering
	\includegraphics[width=0.7\columnwidth]{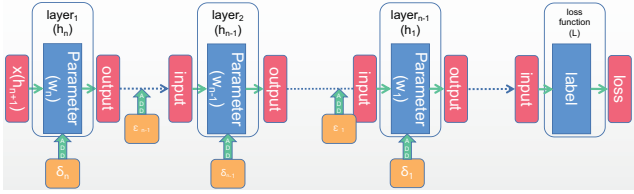}
	\caption{ In current algorithms, like HAWQ, only the parameters are influenced by quantization noise in their assumptions. However, when quantized models are used in the inference process, the outputs of different layers suffer from noise due to the conversion between the different quantization levels of different layers in layerwise quantized models. And the losses on outputs of different layers are more important because the their gradient to losses are non-zero.}
	\label{inference noise}
\end{figure}

Quantization methods for inference are complex. Different algorithms use different assumption to solve the problem. Most of them pay much attention to the noise on parameters in NN\cite{2019HAWQ,2020HAWQV3,gholami2021survey,nagel2020up}. However, in addition to the noise added to the parameters directly, noise is also introduced between different layers in the inference process because different quantization levels or data types of different precisions are used in different layers, which is shown in figure \ref{inference noise}.

After quantization, the quantized loss for a sample, i.e.$\bar{\ell}(\cdot)$,  in the inference process is as follows.
\begin{align*}
	&\bar{\ell}(w,x_i,y_i) = L(h_1(h_2(\cdots h_n(h_{n+1}+\epsilon_n,w_n+\delta_n)\\
	&+\epsilon_{n-1} \cdots,w_2+\delta_2)+\epsilon_1 ,w_1+\delta_1),y_i)
\end{align*}
where $\delta_i$, $i\in{1,\cdots,n}$, and $\epsilon_i$, $i\in[1,...,n]$, are the minor errors that are introduced in model parameter quantization and in data type conversion in the mixed-precision layout scheme, respectively.

Thus, we obtain the following expression based on the basic total differential calculation.
\begin{align}
	\bar{\ell}(w,x_i,y_i) - \ell(w,x_i,y_i) 
	= \sum_{i=1}^n \frac{\partial \ell}{\partial h_{i+1}}  \cdot  \epsilon_i +  \frac{\partial \ell}{\partial w_i}  \cdot \delta_i  \label{orignal inference}
\end{align}
where $\cdot$ is inner product and $*$ is the scalar product in following parts.  For the loss on whole dataset, we can gain

\begin{align}
	\min_{\epsilon \in E }\bar{f}(w) - f(w) &= \frac{1}{m}\sum_{(x_j,y_j)\in \mathbb{D}}	  \sum_{i=1}^n \frac{\partial \ell}{\partial h_{i+1}}  \cdot  \epsilon_i  +  \frac{\partial \ell}{\partial w_i}  \cdot  \delta_i \\
	&=  \frac{1}{m} \sum_{i=1}^n  {\sum_{(x_j,y_j)\in \mathbb{D}}\frac{\partial \ell}{\partial h_{i+1}}} \cdot  \epsilon_i  
	\label{final inference}
\end{align}

where $\bar{f}(w) = \frac{1}{m}\sum \bar{\ell}(\cdot)$.   The reason for second equation in Eq. \ref{final inference} is for a well-trained model, the expectation of $\ell(\cdot)$'s gradient for parameters is zero, i.e., for the $\sum_{(x_j,y_j)\in \mathbb{D}}\frac{\partial \ell}{\partial w}$ components, $\frac{\partial \ell}{\partial w_i} = 0$.

As we can see, current works mixed discussed the quantization for storage and inference \cite{2019HAWQ,2020HAWQV3,dong2019hawq,nahshan2021loss}. Consequently, these works must add a ``fine-tuning'' process, and they still fail in some cases. Moreover, this is why channel-wise quantization methods are booming. In a channel that uses the same data type at all times, the precision loss of the corresponding layer input is usually zero.
\subsubsection{Target and Algorithm Guarantee}

The key is to choose the appropriate $\epsilon$  vector to gain a lower loss model. When the loss of the inner product, i.e., $\sum_{(x_i,y_i) \in \mathbb{D}}\frac{\partial \ell}{\partial h_{i+1}} \cdot\epsilon$, is negative, the loss for the quantized model, i.e., $\bar{f}$, is lower than for the full precision model. An appropriate  $\epsilon$  to produce a negative $\sum_{(x_i,y_i) \in \mathbb{D}}\frac{\partial \ell}{\partial h_{i+1}} \cdot \epsilon$ is our algorithm target.

A frequently asked question is why   $\sum_{(x_j,y_j)\in \mathbb{D}}\frac{\partial \ell}{\partial w}$ is zero but $\sum_{(x_i,y_i) \in \mathbb{D}}\frac{\partial \ell}{\partial h_{i+1}}$ is non-zero. The optimization algorithm is to optimize $w$   in the training process. Thus,  $\sum_{(x_i,y_i) \in \mathbb{D}}\frac{\partial \ell}{\partial h_{i+1}}$ is random in the final model except for the layers with bias terms like the batch norm layer. The bias term will absorb the gradient and train them in the optimization process. What is more, in the model, which mainly consists of identity mapping,  $\sum_{(x_i,y_i) \in \mathbb{D}}\frac{\partial \ell}{\partial h_{i+1}}$  is close to zero vector, and we will show this in the next chapter.

Our problem setting for quantization is different from previous work like HAWQ\cite{2019HAWQ,2020HAWQV3,dong2019hawq,nagel2020up} because these methods do not take the error in the layer's input into consideration, which prevents their work and analysis in the mixed-precision computing area. As a result, these works can only be used to store a compressed neural network on a disk. When the compressed model is stored in memory for inference, these compressed models have to be recovered into the full precision model.

\subsection{ The Map from Mathematical Analysis to Real Engineering and Algorithm Description }

In the above analysis, the whole process is under the condition that $\epsilon$  vector is small enough, which can be used in the total differential method.  However, in practice, the scope of   $\epsilon$  may be within  [-0.1,0.1], which would escape the concept of neighbourhood.     What is more, mapping $\epsilon$ vector into round operation should be fully discussed. This part will show how to deal with the above gap between analysis and engineering.

.

\subsubsection{Round function choice}
We use  the convenient language of probability theory to describe $\frac{\partial \ell}{\partial h_{i+1}}\cdot\epsilon$ for $\epsilon$ is a stochastic vector naturally.  We set $\epsilon = [e_1,e_2,..,e_k]$ and $e_i$ is  i.i.d. random variable. We also set that $\frac{\partial \ell}{\partial h_{i+1}} = [p_1,p_2,...,p_k]$ and $p_i$ is i.i.d. random variable \footnote{We also can treat $p_i$ as the random variable with different distributions or directly use  $\mathbf{E}\frac{\partial \ell}{\partial h_{i+1}}$ vector in following analyses. The   conclusions are the same or close with current analysis.}. $e$ and $p$ are independence to each other.

Then, we have $ \frac{\partial \ell}{\partial h_{i+1}}\cdot\epsilon = \sum_{i=1}^{k}e_i*p_i = kep$ and following Eq. \ref{expection inner product}.
\begin{equation}
	\mathbf{E} \frac{\partial \ell}{\partial h_{i+1}} \cdot \epsilon =  \mathbf{E}  \sum_{i=1}^k e_i*p_i= \mathbf{E}kep = k\mathbf{E}e\mathbf{E}p
	\label{expection inner product}
\end{equation}

For a trained model, the $\mathbf{E}p$ can be computed as  $\mathbf{E}p = \frac{1}{k}*\frac{\partial \ell}{\partial h_{i+1}}\cdot \vec{1}$. Then to gain a negative $\mathbf{E} \frac{\partial \ell}{\partial h_{i+1}}\cdot \epsilon $,  the $\mathbf{E}e$ should be different signs with  $\mathbf{E}p$.

To gain the suitable $\epsilon$  vector, we use the different round functions to ensure the sign of  $\mathbf{E}e$. The roundup function, i.e., the $ceil$ function in python,  will produce an error vector whose all elements are positive. The round down function, i.e., the $floor$ function, will produce an error vector whose all elements are negative. Thus, we are sure that the   $\mathbf{E}e$ is positive and negative by round methods. Although the parameters in layers have strong noise robustness, we still try to add less noise to them. Thus, in the parameters quantization process, we use the rounding method, i.e., the $round$ function in python, to quantize parameters for the rounding method exerts less noise on original data.

\subsubsection{Replace gradient with secant line slope}
Although the elements in the $\epsilon$ vector are not small enough to use the total differential directly, the elements in the $\epsilon$ vector are still small. For example, when using int8 to quantize the res14 model without identity mapping, the element in the $\epsilon$ vector is less than 0.01. The above fact shows that $\|\epsilon_i\|*\|\epsilon_i\|$ is small, which has a tiny influence on the final loss function. Thus, we can use the slope of the secant line to replace the gradient in the total differential.

We define the following  $secant^+(h_i,\Delta),\Delta \in \mathbb{R}^1+$ and  $secant^-(h_i,\Delta),\Delta \in \mathbb{R}^1+$. $\Delta$ is the maximum  error which is introduced by quantization. For example, the  $scale$ parameter in Section 3.1's example is the max error introduced by quantization.

\begin{align*}
	&\bar{\ell_j^{\pm}}(w,x_i,y_i) = L(h_1(h_2(\cdots h_j(h_{j+1}(\cdots)\pm\Delta*\vec{1},w_n)\cdots,w_2),w_1),y_i),\Delta > 0\\
	&\bar{f_j}^\pm(w)=\frac{1}{m}\sum_{x_i,y_i \in Dataset}\bar{\ell^\pm_j}(\cdot), secant(h_i,\Delta)^\pm=\frac{f^\pm -f}{\pm \Delta}
\end{align*}

In the algorithm, we will use $scant^\pm(\cdot)$ to replace $\frac{\partial \ell}{\partial h_{i+1}}^\pm\cdot \vec{1}$.   We use this definition because 1. Compared to computing by the definition of secant, the $secant$ function is easy to be computed. 2. If slope of secant line is $[sec_1,sec_2...,sec_k]$, $secant(h_i,\Delta)^\pm = \mathbf{E}sec$.

Although we know the element in $\epsilon$ vector is less than 0.01 empirically, we still have to set a mechanism in real algorithm design to keep the analysis map into algorithm practice. Thus, we have to set a value $error_{max}$, which $\epsilon$ is small enough for the final loss function. When $\Delta > error_{max}$,  we can choose more bits quantization level or full precision in this layer.

When $\epsilon$ is close to zero, i.e., we use more bits quantization level. For  $\|\frac{\partial \ell}{\partial h_{i+1}}\cdot \epsilon_i \| \leq \| \frac{\partial \ell}{\partial h_{i+1}}\| \| \epsilon_i\|
$,  the performance loss or improvement is small on this layer. So, we directly quantize these parameters and layer's input with this quantization level to reduce computation resources. In algorithm design, we can use $error_{min}$ to control this case.

\subsubsection{The Probability of Getting a Better Model}
%
%
%
%
%
%
%
%
%
%
%
%
%

\begin{algorithm}
	\caption{Radical Mixed-Precision Inference Layout Scheme}	\label{gerenal algorithm}
	\begin{algorithmic}[1]
		\Require Neural network $M$, different quantization levels $[q_1,q_2,...,q_n]$, $error_{min}$,$error_{max}$,$\mu$,calibration dataset $D$
		\State Arrange $Q=[q_1,q_2,...,q_n]$ in ascending order $Q=[q_{i_1},q_{i_2},...,q_{i_n}]$ based on the the size of parameters under $q_i$;//For example Q=[int8,int4,int16] into Q=[int4,int8,int16]	 
	
		\While{$q_i$ in $Q$}
			\While{$Layer_i$ in $M$}
			\If{$Layer_i$ is quantized}
				\State quantize $Layer_i$'s parameters and input by rounding method on $q_i$ quantization level.
			\EndIf
			
			\If{$\Delta>error_{max}$}
				\State continue
			\EndIf
			
			\State Compute $secant^+(h_i,\Delta)$ and $secant^-(h_i,\Delta)$
			
			\If{$\|secant^\pm(\cdot)\|<\mu$}
				\State continue
			\EndIf
			
			\State	$Choice$ = max(max( $secant^-(h_i,\Delta)$,0),min( $secant^+(h_i,\Delta)$,0))
			
			\If{$Choice$ ==  $secant^-(h_i,\Delta)$ }
				\State quantize $Layer_i$'s  input by round down method on $q_i$ quantization level.
			\EndIf
			
			\If{$Choice$ ==  $secant^+(h_i,\Delta)$ }
				\State quantize $Layer_i$'s  input by round up method on $q_i$ quantization level.
			\EndIf
			
			\If{$Choice$ !=  0 }
				\State  quantize $Layer_i$'s parameters by rounding method on $q_i$ quantization level.
			\EndIf

			\EndWhile
		\EndWhile
		
		\State Return	$\bar{M}$
	 
	\end{algorithmic}
\end{algorithm}

To show the probability of getting positive $ \frac{\partial \ell}{\partial h_{i+1}}\cdot\epsilon $, we use chebyshev's theorem, we have following Eq. \ref{chebi}.

\begin{align}
	&P(\frac{\partial \ell}{\partial h_{i+1}} \cdot \epsilon \geq 0)<P(\| \frac{\partial \ell}{\partial h_{i+1}} \cdot \epsilon - \mathbf{E}kep\| \geq  \| \mathbf{E}kep \| ) \notag \\
	&\leq  \frac{Var(kep)}{\|\mathbf{E}kep\|^2} = \frac{Var(e)Var(p)}{\| \mathbf{E}e\mathbf{E}p \|^2 }+\frac{Var(e)}{\| \mathbf{E}e\|^2} + \frac{Var(p)}{\| \mathbf{E}p\|^2}
	\label{chebi}
\end{align}

Based on Eq.\ref{chebi}, we know that to gain a better model performance, for the layer whose  $\| \frac{\partial \ell}{\partial h_{i+1}}\cdot \vec{1}\|$  is large and $Var(p)$ is small, we can use high  quantization level to gain a model which is better than full precision model with high probability. To guarantee the success probability is high, we can set a algorithm parameter $\mu$ and $\sigma$. Algorithm quantize $Layer_i$ only when $\|\mathbf{E}p\|>\mu$ and $Var(p) < \sigma$.

\subsection{Algorithm Description}

Based on the above map between analysis and engineering, we can get algorithm \ref{gerenal algorithm}. Algorithm \ref{gerenal algorithm} is a radical probability algorithm. In algorithm \ref{gerenal algorithm}, we use a high quantization level as a priority to gain a small quantized model. Under the appropriate $\mu$ setting, algorithm \ref{gerenal algorithm} would give a better model with a high probability. Algorithm \ref{gerenal algorithm} does not use the $\sigma$ value to guarantee success because the cost of computing the variance is high.

\section{ Identity Mapping and Quantization}
Although we give a method to gain a better model, we find that ResNets are hard to gain a significant effect. What is more, we find that for most of the quantization algorithms, the performance of ResNets is pretty stable. In this part, we will give theoretical proof of the above phenomena.

The neural network is under the description of the probably approximately correct (PAC) learning framework\cite{Denilson2016Understanding}. A neural network hypothesis class $\mathscr{H}$ consists of the neural networks which share the same structure. The learning algorithms,  $\mathscr{A}$, are SGD and SGD's variants for the neural network hypothesis class. Identity mapping is when the input to some layer is passed directly or as a shortcut to some other layer. The neural networks, which mainly consist of identity mappings, like ResNet or DenseNet, succeed in the CV area. Then, we can gain the following propositions.

\begin{myproposition}
	\label{pp1}
	There is a set of function $\mathscr{G}$. For any random variable vector $x$ and any random variable vector $y$, $\exists g \in \mathscr{G}$ which satisfies $\mathbf{E}g(x)\cdot\mathbf{E}y\leq 0$  and $g(x)$ belongs to $\vec{0}$'s neighborhood. 
	
	For a well-trained neural network $model^*_n\in  \mathscr{H}_n$ by learning algorithm $\mathscr{A}$,  there exists a $model_{n+1}\in  \mathscr{H}_{n+1}$ which is slightly better than $model^*_n$. The difference between  $\mathscr{H}_n$ and $\mathscr{H}_{n+1}$ is the $model_{n+1}\in  \mathscr{H}_{n+1}$ have one more residual block than $model_{n+1}\in  \mathscr{H}_{n}$ and the function in residual block is in $\mathscr{G}$.
\end{myproposition}

\begin{figure}[t]
	\centering
	\includegraphics[width=0.5\linewidth]{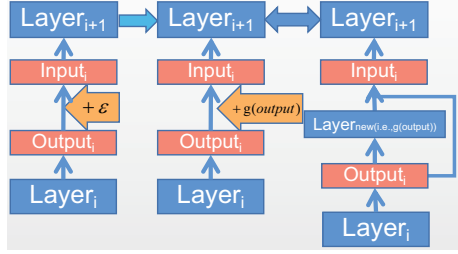}
	\caption{Proof of proposition \ref{pp1}}
	\label{proofpp1}
\end{figure}

Brief proof: From the analysis in algorithm \ref{gerenal algorithm}, we can find an appropriate $\mathbf{E}\epsilon$ that   $\mathbf{E}\frac{\partial \ell}{\partial h_{i+1}}\cdot\epsilon \leq 0$. We can use $g \in  \mathscr{G}$ to replace $\epsilon$. Then, proposition \ref{pp1} is proved, which is also shown in figure \ref{proofpp1}.

\begin{figure}[t]
	\centering
	\includegraphics[width=0.5\linewidth]{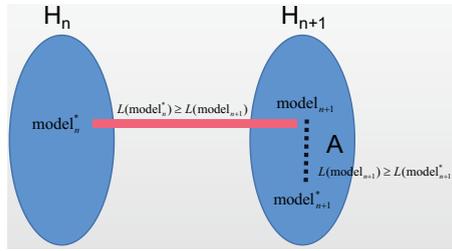}
	\caption{The deeper the neural network is,  the better.}
	\label{fig:side:a}
\end{figure}

\begin{figure}[t]
	\centering
	\includegraphics[width=0.33\linewidth]{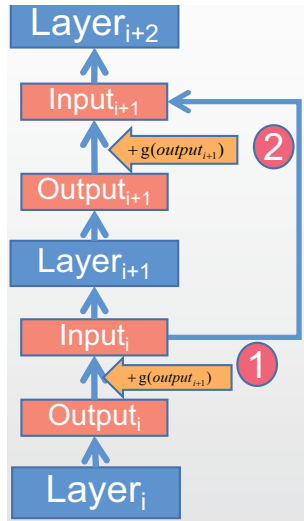}
	\caption{Using the Proof of proposition \ref{pp1} in position 1 repeatedly, we can get ResNet. Using the Proof of proposition \ref{pp1} in position 1 and 2 repeatedly, we can get DenseNet.}
	\label{fig:side:b}
\end{figure}

The set, which consists of $Relu(Conv(\cdot))$, satisfies the requirements of $\mathscr{G}$. Proposition  \ref{pp1} tells us how to structure a deep residual network. Repeatedly using proposition  \ref{pp1}  and retraining the new model would show that for the neural networks consisting of residual blocks like ResNet, the deeper, the better. It is shown in figure \ref{fig:side:a}. Using proposition \ref{pp1} in a different place, we can get different networks, shown in figure \ref{fig:side:b}.

Based on the proposition \ref{pp1}'s structure process, we can prove the following proposition \ref{pp2}.

\begin{myproposition}
	For a dataset's SOTA or close to SOTA residual network, all $\mathbf{E} \frac{\partial \ell}{\partial h_{i+1}}$ are close to zero.
	\label{pp2}
\end{myproposition} 
Brief proof: The SOTA model implies that adding new layers will not improve model performance, i.e., for well-trained  $model^*_n\in  \mathscr{H}_n$ and  well-trained $model^*_{n+1}\in  \mathscr{H}_{n+1}$, $\mathbf{E}_{sample}L(model^*_n,sample) - \mathbf{E}_{sample}L(model^*_{n+1},sample)=0$. So for any $i$ and any appropriate $\mathbf{E}\epsilon$, we have the following Eq \ref{proofzero}.
\begin{align}
	\mathbf{E}\frac{\partial \ell}{\partial h_{i+1}}\cdot\epsilon  
	= \mathbf{E}_{sample}L(model^*_n,sample) - \mathbf{E}_{sample}L(model_{n+1},sample) \notag\\
	<\mathbf{E}_{sample}L(model^*_n,sample) - \mathbf{E}_{sample}L(model^*_{n+1},sample)=0
	\label{proofzero}
\end{align}
Because $i$ and $\mathbf{E}\epsilon$ can be chosen at random, we can tell that $\mathbf{E} \frac{\partial \ell}{\partial h_{i+1}} $ is zero or very close to zero.

Proposition \ref{pp2} shows one of the residual network's SOTA criterion. Then, we can prove the following theorem \ref{pp3}.
\begin{mytheorem}
	The SOTA or near to SOTA residual networks in a dataset exhibit high noise robustness.
	\label{pp3}
\end{mytheorem} 
Brief proof: Based on Eq.\ref{final inference} and proposition \ref{pp2}, we know 
\begin{equation}
	\|\bar{f}(w) - f(w)\|= \| \sum_{i=1}^n  { \mathbf{E}\frac{\partial \ell}{\partial h_{i+1}}} \cdot  \epsilon_i \|\leq   \sum_{i=1}^n  \| \mathbf{E}\frac{\partial \ell}{\partial h_{i+1}}\| \|  \mathbf{E}\epsilon_i\| = 0
\end{equation}

Theorem \ref{pp3} shows that any quantization algorithm can quantize a SOTA or close SOTA residual model with less loss.  Furthermore, Theorem \ref{pp3} also shows that algorithm \ref{gerenal algorithm} cannot improve model performance too much.

\section{Experiment}

In this section, we evaluate the performance of algorithm  \ref{gerenal algorithm}. Our objective is to show that the quantized model gained by algorithm  \ref{gerenal algorithm} is better than the full precision model without "fine-tuning" technology. The improvements are significant, especially for some neural networks that are not SOTA residual networks.

We use CIFAR 10 datasets. The training dataset is split into calibration and training datasets. Furthermore, the size of the calibration dataset is equal to the test dataset.

As discussed in Section 5, the model with many identity mapping structures has strong noise robustness. Thus we have to choose the model which is far from the SOTA model. For CIFAR 10, ResNet 20 is close to the SOTA ResNet model(ResNet 110). In experiments, the performance of ResNet 20 is only higher by 1\% to 2\% in the error rate than ResNet 110. Thus,  we choose ResNet 8 and ResNet 14 as our quantization models. Furtherly, to enlarge the $ \mathbf{E}\frac{\partial \ell}{\partial h_{i+1}}$, we delete the identity mapping structure in our experimental models.  VGG models also are pretty suit for this experiments for the structure of VGG is sequential. However, SOTA VGG model, VGG19, shows the same properties with ResNets--the gradients for middle-output is close to zero. Thus, we use VGG11 and 16 in experiments. 

In quantization practice, quantizing data into INT8 is the most frequently used and ripe choice because current computation devices, like V100GPU, only support INT8, INT16 and INT32 computing in hardware. Thus, we use mixed-precision INT8 and full-precision(FP32) in our experiments.

In Cifar 10 experiment, the min and max value for a quantization vector is decided by whole calibration dataset. Before quantization process, we will process whole calibration dataset in full precise and find the min and max value and compute $\Delta$. We use this setting because we want to enlarge the noise and get a obvious experimental results. However, in VGG experiments, under this setting,  we cannot find an appropriate layer to quantize because all $\Delta$s are large or $Choice$s variable are zero. Thus, we use the min/max on current quantization vector like HAWQ's experiments to compute $\Delta$.

ResNet8 has a total of 23 layers, and Res14 has a total of 41 layers. $\Delta$, which is the error caused by the INT8 quantization level for layers' input, is under 0.01 for most of the elements, but a few of them are above this range.  So, we use 0.01 in all $secant(\cdot)$ functions.  To gain a high performance model as radically as possible, we set $error_{min} = 0$, $\mu = 0$ and $error_{max} = 0.1$.  Because we use conservation plan in compute $\Delta$ in VGG experiments, Thus, we also use the same setting with Resnet experiments.

The loss for ResNet8 with the full precision is 0.3896, and the loss for the quantization model is 0.3782 with all layers quantized.   The loss for ResNet14 with full precision is 0.3634, and the loss for the quantization model is 0.3576 with ten layers quantized. Only ten layers are quantized because the $choice$ variable in algorithm \ref{gerenal algorithm} is zero for most of the layers.  To quantize ResNet14 further, we have to use another quantization level to gain a better performance model, like the INT16 level. However, the current low precision computation toolchains are uncompleted. For example, CUDNN only supports FP32, FP16, INT32 and INT8 input. Further ResNet14 experiments on different fabricated quantization settings are nonsense. The loss for VGG11 with the full precision is 0.0019, and the loss for the quantization model is 0.0017 with all layers quantized.   The loss for VGG16 with full precision is 0.00908., and the loss for the quantization model is 0.0086 with all layers quantized. 

\section{Conclusion}

This paper shows that quantization technology can improve the model's performance, i.e., gain a lower loss. Moreover, based on our analysis, we propose a Radical Mixed-Precision Inference Layout Scheme, which could produce a quantized model which is better than the full-precision model. We also show that residual networks are very resistant to noise. This means that the performance of a SOTA residual network is stable for any quantization algorithms.

In this paper, we also want to show the idea that the neural networks should be categorized into more specific classes because they share the different mathematical properties. The same algorithm shows great different performance on different models and datasets. In algorithm research, we also should show that under which condition, the algorithm performance which current researches ignore.

 

\bibliography{mybibtex}


\begin{thebibliography}{21}
\ifx \bisbn   \undefined \def \bisbn  #1{ISBN #1}\fi
\ifx \binits  \undefined \def \binits#1{#1}\fi
\ifx \bauthor  \undefined \def \bauthor#1{#1}\fi
\ifx \batitle  \undefined \def \batitle#1{#1}\fi
\ifx \bjtitle  \undefined \def \bjtitle#1{#1}\fi
\ifx \bvolume  \undefined \def \bvolume#1{\textbf{#1}}\fi
\ifx \byear  \undefined \def \byear#1{#1}\fi
\ifx \bissue  \undefined \def \bissue#1{#1}\fi
\ifx \bfpage  \undefined \def \bfpage#1{#1}\fi
\ifx \blpage  \undefined \def \blpage #1{#1}\fi
\ifx \burl  \undefined \def \burl#1{\textsf{#1}}\fi
\ifx \doiurl  \undefined \def \doiurl#1{\url{https://doi.org/#1}}\fi
\ifx \betal  \undefined \def \betal{\textit{et al.}}\fi
\ifx \binstitute  \undefined \def \binstitute#1{#1}\fi
\ifx \binstitutionaled  \undefined \def \binstitutionaled#1{#1}\fi
\ifx \bctitle  \undefined \def \bctitle#1{#1}\fi
\ifx \beditor  \undefined \def \beditor#1{#1}\fi
\ifx \bpublisher  \undefined \def \bpublisher#1{#1}\fi
\ifx \bbtitle  \undefined \def \bbtitle#1{#1}\fi
\ifx \bedition  \undefined \def \bedition#1{#1}\fi
\ifx \bseriesno  \undefined \def \bseriesno#1{#1}\fi
\ifx \blocation  \undefined \def \blocation#1{#1}\fi
\ifx \bsertitle  \undefined \def \bsertitle#1{#1}\fi
\ifx \bsnm \undefined \def \bsnm#1{#1}\fi
\ifx \bsuffix \undefined \def \bsuffix#1{#1}\fi
\ifx \bparticle \undefined \def \bparticle#1{#1}\fi
\ifx \barticle \undefined \def \barticle#1{#1}\fi
\bibcommenthead
\ifx \bconfdate \undefined \def \bconfdate #1{#1}\fi
\ifx \botherref \undefined \def \botherref #1{#1}\fi
\ifx \url \undefined \def \url#1{\textsf{#1}}\fi
\ifx \bchapter \undefined \def \bchapter#1{#1}\fi
\ifx \bbook \undefined \def \bbook#1{#1}\fi
\ifx \bcomment \undefined \def \bcomment#1{#1}\fi
\ifx \oauthor \undefined \def \oauthor#1{#1}\fi
\ifx \citeauthoryear \undefined \def \citeauthoryear#1{#1}\fi
\ifx \endbibitem  \undefined \def \endbibitem {}\fi
\ifx \bconflocation  \undefined \def \bconflocation#1{#1}\fi
\ifx \arxivurl  \undefined \def \arxivurl#1{\textsf{#1}}\fi
\csname PreBibitemsHook\endcsname

\bibitem{han2015learning}
\begin{botherref}
\oauthor{\bsnm{Han}, \binits{S.}},
\oauthor{\bsnm{Pool}, \binits{J.}},
\oauthor{\bsnm{Tran}, \binits{J.}},
\oauthor{\bsnm{Dally}, \binits{W.J.}}:
Learning both weights and connections for efficient neural networks.
arXiv preprint arXiv:1506.02626
(2015)
\end{botherref}
\endbibitem

\bibitem{li2016pruning}
\begin{botherref}
\oauthor{\bsnm{Li}, \binits{H.}},
\oauthor{\bsnm{Kadav}, \binits{A.}},
\oauthor{\bsnm{Durdanovic}, \binits{I.}},
\oauthor{\bsnm{Samet}, \binits{H.}},
\oauthor{\bsnm{Graf}, \binits{H.P.}}:
Pruning filters for efficient convnets.
arXiv preprint arXiv:1608.08710
(2016)
\end{botherref}
\endbibitem

\bibitem{mao2017exploring}
\begin{botherref}
\oauthor{\bsnm{Mao}, \binits{H.}},
\oauthor{\bsnm{Han}, \binits{S.}},
\oauthor{\bsnm{Pool}, \binits{J.}},
\oauthor{\bsnm{Li}, \binits{W.}},
\oauthor{\bsnm{Liu}, \binits{X.}},
\oauthor{\bsnm{Wang}, \binits{Y.}},
\oauthor{\bsnm{Dally}, \binits{W.J.}}:
Exploring the regularity of sparse structure in convolutional neural networks.
arXiv preprint arXiv:1705.08922
(2017)
\end{botherref}
\endbibitem

\bibitem{hinton2015distilling}
\begin{botherref}
\oauthor{\bsnm{Hinton}, \binits{G.}},
\oauthor{\bsnm{Vinyals}, \binits{O.}},
\oauthor{\bsnm{Dean}, \binits{J.}}:
Distilling the knowledge in a neural network.
arXiv preprint arXiv:1503.02531
(2015)
\end{botherref}
\endbibitem

\bibitem{ullrich2017soft}
\begin{botherref}
\oauthor{\bsnm{Ullrich}, \binits{K.}},
\oauthor{\bsnm{Meeds}, \binits{E.}},
\oauthor{\bsnm{Welling}, \binits{M.}}:
Soft weight-sharing for neural network compression.
arXiv preprint arXiv:1702.04008
(2017)
\end{botherref}
\endbibitem

\bibitem{2019Fully}
\begin{bchapter}
\bauthor{\bsnm{Li}, \binits{R.}},
\bauthor{\bsnm{Wang}, \binits{Y.}},
\bauthor{\bsnm{Liang}, \binits{F.}},
\bauthor{\bsnm{Qin}, \binits{H.}},
\bauthor{\bsnm{Fan}, \binits{R.}}:
\bctitle{Fully quantized network for object detection}.
In: \bbtitle{2019 IEEE/CVF Conference on Computer Vision and Pattern
  Recognition (CVPR)}
(\byear{2019})
\end{bchapter}
\endbibitem

\bibitem{2020Channel}
\begin{botherref}
\oauthor{\bsnm{Qian}, \binits{X.}},
\oauthor{\bsnm{Li}, \binits{V.}},
\oauthor{\bsnm{Darren}, \binits{C.}}:
Channel-wise hessian aware trace-weighted quantization of neural networks
(2020)
\end{botherref}
\endbibitem

\bibitem{dong2019hawq}
\begin{botherref}
\oauthor{\bsnm{Dong}, \binits{Z.}},
\oauthor{\bsnm{Yao}, \binits{Z.}},
\oauthor{\bsnm{Cai}, \binits{Y.}},
\oauthor{\bsnm{Arfeen}, \binits{D.}},
\oauthor{\bsnm{Gholami}, \binits{A.}},
\oauthor{\bsnm{Mahoney}, \binits{M.W.}},
\oauthor{\bsnm{Keutzer}, \binits{K.}}:
Hawq-v2: Hessian aware trace-weighted quantization of neural networks.
arXiv preprint arXiv:1911.03852
(2019)
\end{botherref}
\endbibitem

\bibitem{2019HAWQ}
\begin{botherref}
\oauthor{\bsnm{Dong}, \binits{Z.}},
\oauthor{\bsnm{Yao}, \binits{Z.}},
\oauthor{\bsnm{Gholami}, \binits{A.}},
\oauthor{\bsnm{Mahoney}, \binits{M.}},
\oauthor{\bsnm{Keutzer}, \binits{K.}}:
Hawq: Hessian aware quantization of neural networks with mixed-precision.
IEEE
(2019)
\end{botherref}
\endbibitem

\bibitem{morgan1991experimental}
\begin{bchapter}
\bauthor{\bsnm{Morgan}, \binits{N.}}, \betal:
\bctitle{Experimental determination of precision requirements for
  back-propagation training of artificial neural networks}.
In: \bbtitle{Proc. Second Int’l. Conf. Microelectronics for Neural Networks},
pp. \bfpage{9}--\blpage{16}
(\byear{1991}).
\bcomment{Citeseer}
\end{bchapter}
\endbibitem

\bibitem{courbariaux2015binaryconnect}
\begin{bchapter}
\bauthor{\bsnm{Courbariaux}, \binits{M.}},
\bauthor{\bsnm{Bengio}, \binits{Y.}},
\bauthor{\bsnm{David}, \binits{J.-P.}}:
\bctitle{Binaryconnect: Training deep neural networks with binary weights
  during propagations}.
In: \bbtitle{Advances in Neural Information Processing Systems},
pp. \bfpage{3123}--\blpage{3131}
(\byear{2015})
\end{bchapter}
\endbibitem

\bibitem{2020HAWQV3}
\begin{botherref}
\oauthor{\bsnm{Yao}, \binits{Z.}},
\oauthor{\bsnm{Dong}, \binits{Z.}},
\oauthor{\bsnm{Zheng}, \binits{Z.}},
\oauthor{\bsnm{Gholami}, \binits{A.}},
\oauthor{\bsnm{Keutzer}, \binits{K.}}:
Hawqv3: Dyadic neural network quantization
(2020)
\end{botherref}
\endbibitem

\bibitem{gholami2021survey}
\begin{botherref}
\oauthor{\bsnm{Gholami}, \binits{A.}},
\oauthor{\bsnm{Kim}, \binits{S.}},
\oauthor{\bsnm{Dong}, \binits{Z.}},
\oauthor{\bsnm{Yao}, \binits{Z.}},
\oauthor{\bsnm{Mahoney}, \binits{M.W.}},
\oauthor{\bsnm{Keutzer}, \binits{K.}}:
A survey of quantization methods for efficient neural network inference.
arXiv preprint arXiv:2103.13630
(2021)
\end{botherref}
\endbibitem

\bibitem{wu2018mixed}
\begin{botherref}
\oauthor{\bsnm{Wu}, \binits{B.}},
\oauthor{\bsnm{Wang}, \binits{Y.}},
\oauthor{\bsnm{Zhang}, \binits{P.}},
\oauthor{\bsnm{Tian}, \binits{Y.}},
\oauthor{\bsnm{Vajda}, \binits{P.}},
\oauthor{\bsnm{Keutzer}, \binits{K.}}:
Mixed precision quantization of convnets via differentiable neural architecture
  search.
arXiv preprint arXiv:1812.00090
(2018)
\end{botherref}
\endbibitem

\bibitem{wang2019haq}
\begin{bchapter}
\bauthor{\bsnm{Wang}, \binits{K.}},
\bauthor{\bsnm{Liu}, \binits{Z.}},
\bauthor{\bsnm{Lin}, \binits{Y.}},
\bauthor{\bsnm{Lin}, \binits{J.}},
\bauthor{\bsnm{Han}, \binits{S.}}:
\bctitle{Haq: Hardware-aware automated quantization with mixed precision}.
In: \bbtitle{Proceedings of the IEEE/CVF Conference on Computer Vision and
  Pattern Recognition},
pp. \bfpage{8612}--\blpage{8620}
(\byear{2019})
\end{bchapter}
\endbibitem

\bibitem{yu2020search}
\begin{bchapter}
\bauthor{\bsnm{Yu}, \binits{H.}},
\bauthor{\bsnm{Han}, \binits{Q.}},
\bauthor{\bsnm{Li}, \binits{J.}},
\bauthor{\bsnm{Shi}, \binits{J.}},
\bauthor{\bsnm{Cheng}, \binits{G.}},
\bauthor{\bsnm{Fan}, \binits{B.}}:
\bctitle{Search what you want: Barrier panelty nas for mixed precision
  quantization}.
In: \bbtitle{European Conference on Computer Vision},
pp. \bfpage{1}--\blpage{16}
(\byear{2020}).
\bcomment{Springer}
\end{bchapter}
\endbibitem

\bibitem{2020Effective}
\begin{bchapter}
\bauthor{\bsnm{Demidovskij}, \binits{A.}},
\bauthor{\bsnm{Smirnov}, \binits{E.}}:
\bctitle{Effective post-training quantization of neural networks for inference
  on low power neural accelerator}.
In: \bbtitle{2020 International Joint Conference on Neural Networks (IJCNN)}
(\byear{2020})
\end{bchapter}
\endbibitem

\bibitem{Denilson2016Understanding}
\begin{barticle}
\bauthor{\bsnm{Denilson}},
\bauthor{\bsnm{Barbosa}}:
\batitle{Understanding machine learning: from theory to algorithms}.
\bjtitle{Computing reviews}
\bvolume{57}(\bissue{4}),
\bfpage{238}--\blpage{238}
(\byear{2016})
\end{barticle}
\endbibitem

\bibitem{2019Relay}
\begin{botherref}
\oauthor{\bsnm{Roesch}, \binits{J.}},
\oauthor{\bsnm{Lyubomirsky}, \binits{S.}},
\oauthor{\bsnm{Kirisame}, \binits{M.}},
\oauthor{\bsnm{Pollock}, \binits{J.}},
\oauthor{\bsnm{Tatlock}, \binits{Z.}}:
Relay: A high-level ir for deep learning
(2019)
\end{botherref}
\endbibitem

\bibitem{nagel2020up}
\begin{bchapter}
\bauthor{\bsnm{Nagel}, \binits{M.}},
\bauthor{\bsnm{Amjad}, \binits{R.A.}},
\bauthor{\bsnm{Van~Baalen}, \binits{M.}},
\bauthor{\bsnm{Louizos}, \binits{C.}},
\bauthor{\bsnm{Blankevoort}, \binits{T.}}:
\bctitle{Up or down? adaptive rounding for post-training quantization}.
In: \bbtitle{International Conference on Machine Learning},
pp. \bfpage{7197}--\blpage{7206}
(\byear{2020}).
\bcomment{PMLR}
\end{bchapter}
\endbibitem

\bibitem{nahshan2021loss}
\begin{barticle}
\bauthor{\bsnm{Nahshan}, \binits{Y.}},
\bauthor{\bsnm{Chmiel}, \binits{B.}},
\bauthor{\bsnm{Baskin}, \binits{C.}},
\bauthor{\bsnm{Zheltonozhskii}, \binits{E.}},
\bauthor{\bsnm{Banner}, \binits{R.}},
\bauthor{\bsnm{Bronstein}, \binits{A.M.}},
\bauthor{\bsnm{Mendelson}, \binits{A.}}:
\batitle{Loss aware post-training quantization}.
\bjtitle{Machine Learning}
\bvolume{110}(\bissue{11}),
\bfpage{3245}--\blpage{3262}
(\byear{2021})
\end{barticle}
\endbibitem

\end{thebibliography}

\end{document}